\documentclass{article}
\usepackage{arxiv}
\usepackage{amsfonts}
\usepackage{amsmath}
\usepackage{amssymb}
\usepackage{bm}
\usepackage{booktabs}
\usepackage{dsfont}
\usepackage{graphicx}
\usepackage{subfig}
\usepackage{epstopdf}
\usepackage{epsfig}
\usepackage{url}
\usepackage{color}
\usepackage{array}
\usepackage{threeparttable}
\usepackage{lineno}
\usepackage{multirow}
\usepackage{hyperref}
\usepackage{gensymb}
\usepackage[ruled]{algorithm2e}
\usepackage{cite}
\usepackage{textcomp}

\title{Human Activity Recognition based on Dynamic Spatio-Temporal Relations}

\author{ Zhenyu~Liu \\
  Department of Science and Technology Teaching\\
  China University of Political Science and Law\\
  Beijing 100088\\
  \texttt{lzhy@cupl.edu.cn} \\
   \And
 Yaqiang~Yao \\
  School of Computer and Science Technology\\
  University of Science and Technique of China\\
  Hefei, Anhui 230027\\
  \texttt{yaoyaq@ustc.edu.cn} \\
  \And
 Yan~Liu \\
  School of Computer and Science Technology\\
  University of Science and Technique of China\\
  Hefei, Anhui 230027\\
  \texttt{ready@mail.ustc.edu.cn} \\
  \And
 Yuening Zhu \\
  School of Materials and Engineering\\
  Shanghai University\\
  Shanghai, China\\
  \texttt{1209807881@qq.com} \\
  \And
  Zhenchao Tao \\
  The First Affiliated Hospital \\
  University of Science and Technique of China\\
  Hefei, Anhui 230027\\
  \texttt{taozhenchao@sina.com} \\
  \And
  Lei Wang \\
  The Third People's Hospital of Hefei\\
  Hefei, Anhui 230000\\
  \And
   Yuhong Feng \\
  School of Computer and Software\\
  Shenzhen University\\
  Shenzhen, China\\
  \texttt{yuhongf@szu.edu.cn} \\
}

\begin{document}
\maketitle

\begin{abstract}
Human activity, which usually consists of several actions (sub-activities), generally covers interactions among persons and/or objects. In particular, human actions involve certain spatial and temporal relationships, are the components of more complicated activity, and evolve dynamically over time. Therefore, the description of a single human action and the modeling of the evolution of successive human actions are two major issues in human activity recognition. In this paper, we develop a method for human activity recognition that tackles these two issues. In the proposed method, an activity is divided into several successive actions represented by spatio–temporal patterns, and the evolution of these actions are captured by a sequential model. A refined comprehensive spatio–temporal graph is utilized to represent a single action, which is a qualitative representation of a human action incorporating both the spatial and temporal relations of the participant objects. Next, a discrete hidden Markov model is applied to model the evolution of action sequences. Moreover, a fully automatic partition method is proposed to divide a long-term human activity video into several human actions based on variational objects and qualitative spatial relations. Finally, a hierarchical decomposition of the human body is introduced to obtain a discriminative representation for a single action. Experimental results on the Cornell Activity Dataset demonstrate the efficiency and effectiveness of the proposed approach, which will enable long videos of human activity to be better recognized.
\end{abstract}


\section{Introduction}
\label{sec:introduction}

The automated recognition of human behavior in a video has attracted much interest in the computer-vision domain \cite{aggarwal2011human,chaquet2013survey,hong2019variant,yao2018human,hong2018disturbance} because of its wide range of applications, including video surveillance
\cite{lao2009automatic}, health care and social assistance \cite{jalal2012depth}, human-computer interactions, entertainment\cite{fothergill2012instructing}, and so on. Previous work on human action recognition mainly focuses on the extraction of local space–time features such as SIFT \cite{lowe2004distinctive}, HOG/HOF \cite{laptev2008learning}, and MBH \cite{wang2013dense} from two-dimensional (2D) frame images. Recently, with the emergence of low-cost depth sensors, depth information has been appended in quite a few  approaches \cite{hussein2013human,wang2014learning,xia2012view} and improves the recognition performance substantially. However, these approaches only pay attention to the low-level features of the frame image and ignore the interactions of the participating components, which restricts the recognition of these methods to simple human activity.

Human activity often involves interactions of one or more persons and/or objects. For example, “two people playing” is a person–person interaction, and “a person picking up a ball” is a human–object interaction. These interactions would suffer from serious ambiguity using a low-level feature representation. Besides, one activity comprises several simple sub-activities. In the following, we refer to these sub-activities as actions. As an example, the human activity “microwaving food” requires a person to perform three actions: 1) open the microwave oven door, 2) put the food in, and 3) close the microwave oven door. The task of human activity recognition is usually decomposed into two major topics. The first is human action representation, and the second is the modeling of the evolution of successive human action sequences in a complicated activity. This paper addresses both issues.

Human actions consist of spatio–temporal patterns. For any human action performed with several objects, a spatial relation exists between any two objects in each frame image. Temporal relations are generated from the relative pairwise spatial interactions of different segments. Because the human body is an articulated system of rigid segments connected by joints, these segments are regarded as objects. For example, “hands up” is a human action mainly performed by three parts of the human body: the left/right hands, the hip, and the head. At the start of action, both hands are beside the hip and below the head. There are two spatial relations, “both hands beside the hip” and “both hands below the head.” Then, both hands are above the hip and beside the head. Two new spatial relations “both hands above the hip” and “both hands beside the head” appear for a moment. Finally, the hands are above both the hip and the head. The spatial relation “both hands beside the hip” remains as the third spatial relation “both hands above the head” occurs. In the meantime, temporal relations are applied to organize these spatial relations. However, for any human activity composed of several action sequences, each component may occur over various timescales, which makes the recognition task more challenging.

Taking the above problems into consideration, we propose an efficient method to recognize human activity in this paper based on the qualitative spatio–temporal graph proposed in \cite{sridhar2010unsupervised}. In particular, a qualitative spatio–temporal graph is a comprehensive description of the interactions of objects participating in a human action. Both qualitative spatial and temporal relations are well organized in this graph. However, spatial relations only consider the distance relation between objects and ignore the direction relation, which discards important information. We hence incorporate the direction relation into the spatial–temporal graph to improve the representability. In contrast, to distinguish different graphs, a hierarchical method is utilized to decompose the human body into different parts to construct several qualitative spatio–temporal graphs, which are concatenated to build the representation of the whole human action graph. Then, the derived qualitative spatio–temporal graph is disassembled into cell graphs to better measure the features. The histogram statistics are employed to convert the qualitative spatio–temporal graph into a computable vector, which is utilized to represent potential human actions in the subsequent procedures.

\begin{figure}
\centering
\includegraphics[width=0.9\linewidth]{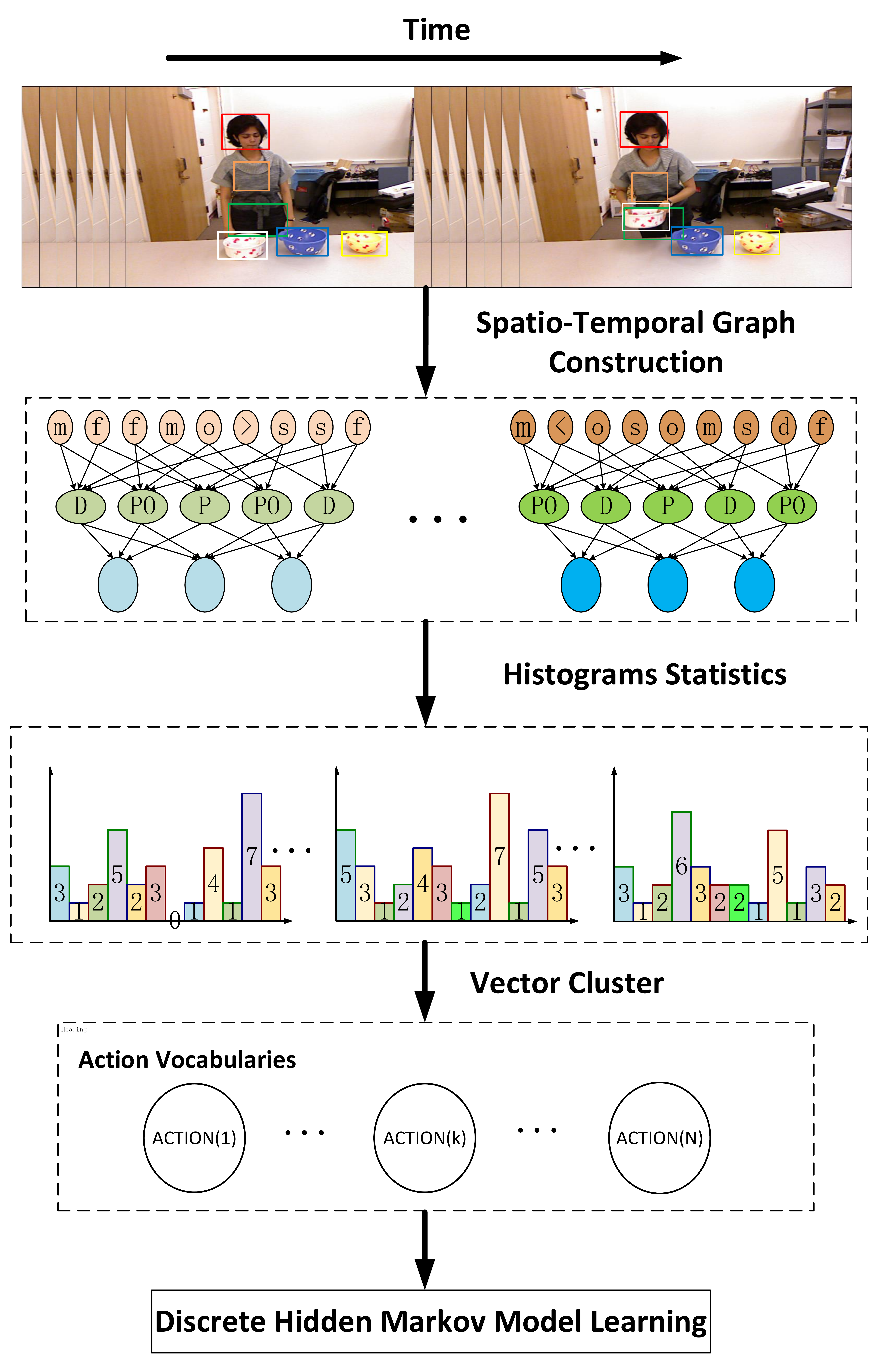}
\caption{General framework of the proposed approach.}
\label{fig:procedure}
\end{figure}

In the proposed approach, discrete Hidden Markov Models (HMMs) \cite{rabiner1989tutorial}  are employed to model the dynamic qualitative spatio-–temporal graphs, which describe the evolution of human action sequences. Specifically, a human activity video is automatically divided into segments when a significant variation of in spatial relations occurs between any pairwise objects. We construct the corresponding qualitative spatio-–temporal graphs based on the specific number of successive sequence segments with constant steps. Next, the distance between qualitative spatio-–temporal graphs is measured to cluster the derived graphs into a vocabulary of $k$ actions, where each word in the vocabulary represents a distinct potential action. Finally, discrete HMMs are applied to recognize human activities. The same procedures are employed in training and recognition phrase. Fig.~\ref{fig:procedure} illustrates the general framework of the proposed approach.

From a short-term perspective, the interactions of the objects are represented by a distinct and robust qualitative spatial–temporal graph. The evolution of qualitative spatio–temporal graphs displays the dynamic variations of human action sequences over a long time period. In this work, we represent a long-term human activity by dynamic qualitative spatial–temporal graphs constructed over a short time period, while a bag-of-words approach is applied to distinguish distinctive simple graphs. We then map the short-duration graphs into discrete symbols to train a discrete HMM for each activity. In contrast to the work \cite{koppula2013learningb} that adopts a Markov random field to jointly model sub-activities and object affordances, our work highlights the interaction of participating objects from both spatial and temporal relations, and models the evolution of a human activity from both microcosmic and macroscopic perspectives. Experimental results on the Cornell Activity Dataset (CAD-120) demonstrate the effectiveness of the proposed approach. CAD-120 is a publicly available benchmark dataset, and includes video sequences of interactions among objects and humans performing daily real-world activities. These long human activity videos (high-level) can be considered as concatenations of actions (low-level) and are ideally compatible with the proposed methodology.

The main contribution of this paper includes the following three aspects:
\begin{enumerate}
\item We develop an efficient method for human activity recognition, in which a long-term human activity (high-level) video is divided into several successive human actions (low-level) represented as spatio-–temporal patterns, and the evolution of these human actions are modeled by HMMs.
\item We improve the qualitative spatio-temporal graph presented in \cite{sridhar2010unsupervised} with direction relations to represent human actions. This graph distinctively describes action and is a robust representation when both ground-truth and automatically extracted track data are used.
\item We apply a fully automatic and efficient tool to partition a human activity video into several successive human action video sequences. Moreover, a hierarchical decomposition of human body parts for component action representation is utilized to improve the proposed approach.
\end{enumerate}

The remainder of this paper is organized as follows. A brief review of the related work on human behavior recognition is stated in Section \ref{sec:relatedwork}. The construction of the improved qualitative spatio-temporal graph is described in Section \ref{sec:graph}. Section \ref{sec:representation} presents the human action feature representation, including the capture of dynamic patterns with a sliding window, hierarchical decomposition of the human body, and histogram statistics of the spatio–temporal graph. Section \ref{sec:recognition} details the transformation from derived features to visual words and the application of discrete HMMs to activity training and recognition. The implementation details on the CAD120 benchmark, experimental results, and ablation analysis are presented in Section \ref{sec:experiment}. Finally, we conclude our work in Section \ref{sec:conclusion}.

\section{Related Work}
\label{sec:relatedwork}
Human activities consist of sub-level activities (actions), which form the temporal evolution of spatial patterns. How to  effectively and efficiently model the spatio–temporal patterns of human behavior has attracted extensive investigations. For example, the work in \cite{lowe2004distinctive,scovanner20073} consider the human action as in the 3D volume where the 2D SIFT descriptor is concatenated to the time dimension. Wang et al. \cite{wang2013dense} found that incorporating motion boundary histograms (MBH \cite{dalal2006human}) with dense trajectories yielded better results on a large number of human action datasets. Yuan et al. \cite{yuan2010middle} defined the SIFT trajectories with consistent spatial structure and consistent motion as middle-level components, and model the spatial and temporal relationships between components, respectively, with discrete states. Xu et al. \cite{xu2017learning,xu2017hierarchical} modeled the spatio–temporal pattern as the a two-layer hidden conditional random field. Gong et al. proposed a general framework, in which the original signals are first transformed to recurrent models, and the spatio–temporal patterns are learned in the model space \cite{li2018symbolic,gong2018sequential,chen2014cognitive}. The learning in the model space transforms the original series to a recurrent neural networks (RNN), tries to calculate the `distance' between RNNs, and conducts the learning in the RNN space \cite{ChenTRY14,chen2013model,gong2016model}. The representation and discrimination abilities have been investigated later \cite{chen2015model} and the multi-objective version has been proposed \cite{gong2018multiobjective}. Compared with this proposed method, these methods are lack of the modeling of interactions between the human body and participating objects, which is critical for complicated human activity recognition.

As a human body can be viewed as the articulated system of rigid segments connected by joints, human actions can be abstracted as the continuous evolution of spatial configurations of such systems (i.e., body postures) \cite{knutzen1998kinematics}. With the emergence of Kinect that facilitates the exaction of joint patterns with depth images, the so-called skeleton-based approaches have turned out to be more applicable for human action recognitions \cite{yang2014super,du2015hierarchical}. For example, Xia et al. \cite{xia2012view} presented a novel descriptor for the compact representation of postures called histograms of 3D joint locations, which uses human joint positions for joint-location binning in a modified spherical coordinate system. Yao et al. \cite{yao2018human} combined the temporal relation of compact action snippets with manifold learning to propose an efficient posture tendency descriptor that concatenates posture tendency descriptors in hierarchical and temporal order. In contrast, our proposed method represents human actions based on the qualitative descriptions of the spatio–temporal relations of human skeleton joints, which can save an enormous amount of computation.

The approaches of qualitative description of patio–temporal relations of objects can be dated back to Region Connection Calculus \cite{li2003region} and Allen's Interval Algebra \cite{allen1983maintaining}, respectively. These relations are combined using a qualitative spatio-temporal graph \cite{sridhar2010unsupervised} to represent an activity. In contrast, Tayyub et al. \cite{tayyub2014qualitative} assembled the statistics of qualitative spatial, qualitative temporal, and quantitative spatial features, which was followed by an automatic feature selection for human activity recognition. The decent performance of the method was ascribed to the quantitative spatial information. However, this method works poorly with long-term motions because of the discard of the temporal information of human actions.

As neural-based approaches have achieved impressive successes in many visual tasks, the models such as the Recurrent Neural Network (RNN) and the Long-Short Term Memory (LSTM) are increasingly applied for modeling human behavior \cite{donahue2015long,du2015hierarchical,mahasseni2016regularizing}. For instance, Du et al. \cite{du2015hierarchical} provided a hierarchical recurrent neural network for human action recognition, in which human actions are represented by the trajectories of the skeleton joints. It is worth to note that, compared with other temporal dynamics modeling approaches, the LSTM model features simultaneous learning of spatial features and dependency information over time. However, the major bottlenecks of neural models are still the requirement of large numbers of samples and the computationally expensive training costs.

As mentioned above, complicated human actions are often the result of interactions between the (parts of) human body and participating objects. Explicit representation of interactions have a promising potential to learn spatio–temporal patterns more efficiently than those representing scenes with global descriptors. Herzig et al. \cite{herzig2019spatio} utilized the Faster R-CNN to detect the bounding boxes of objects and relations between each box to capture the interactions between objects, and subsequently aggregated the features of the the bounding boxes and relations into two disentangled spatial and temporal context features for classification. Tang et al. \cite{tang2019coherence} presented a spatio-temporal context coherence constraint and a global context coherence constraint to distinguish the relevant motions and quantify their contributions to the group activity, respectively, which suppressing the irrelevant motions to address the group activity recognition problem. Lu et al. \cite{lu2019gaim} proposed a graph attention interaction model to address the unbalanced interaction relationship problem at personal and group levels in collective activity recognition. In contrast to these approaches, our proposed method captures interactions by the changes in qualitative spatial and temporal relations of skeleton joints and participating objects, which results in robust performance with respect to both ground truth and noisy data.

\section{Qualitative Spatio-Temporal Graph}
\label{sec:graph}
Human actions are spatio-temporal patterns with interactions among human body joints and/or objects. In the proposed approach, we extend the  qualitative spatio-temporal graph \cite{sridhar2010unsupervised} with direction relations to better represent a human action. In this section, we first present the basic definitions of the primary elements in the improved qualitative spatio-temporal graph, and then describe the process of constructing a qualitative spatial-temporal graph.

\begin{figure*}[t]
\centering
\subfloat[]{\includegraphics[width=0.3\linewidth]{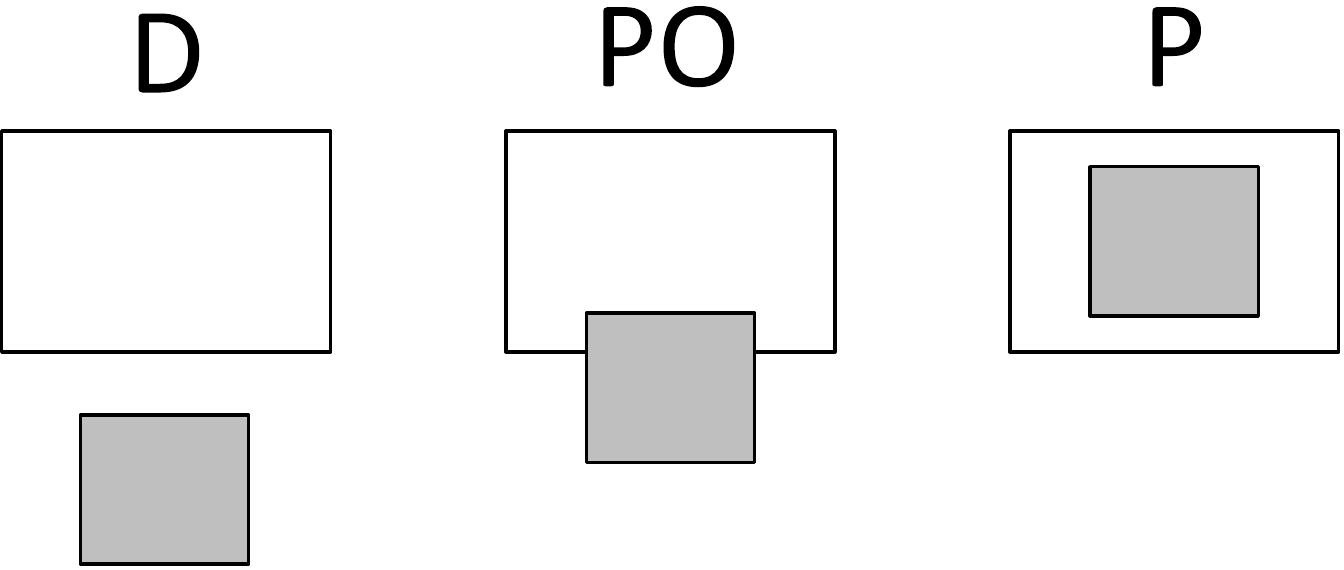}\label{subfig:distanceRelation}}
\qquad
\subfloat[]{\includegraphics[width=0.6\linewidth]{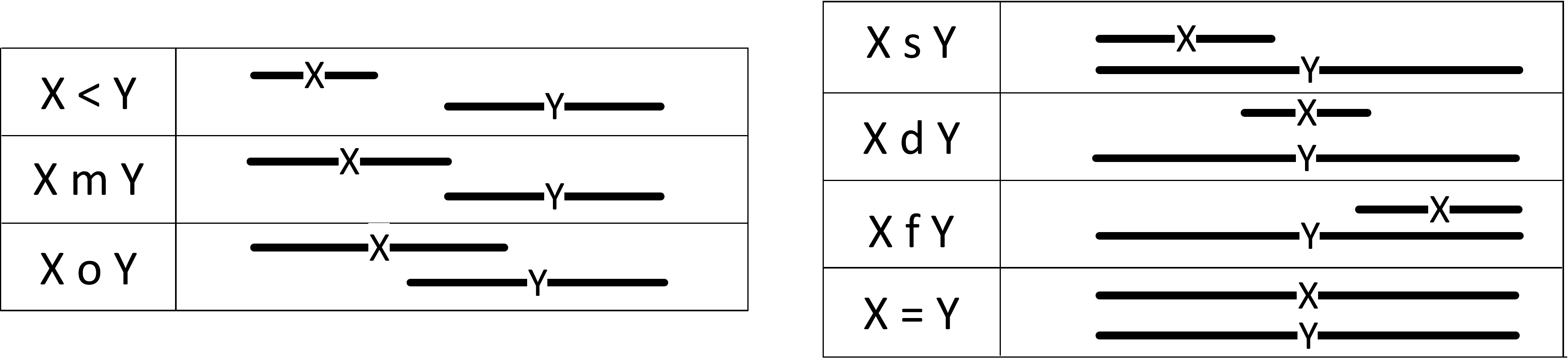}\label{subfig:temporalRelation}} \\
\subfloat[]{\includegraphics[width=0.18\linewidth]{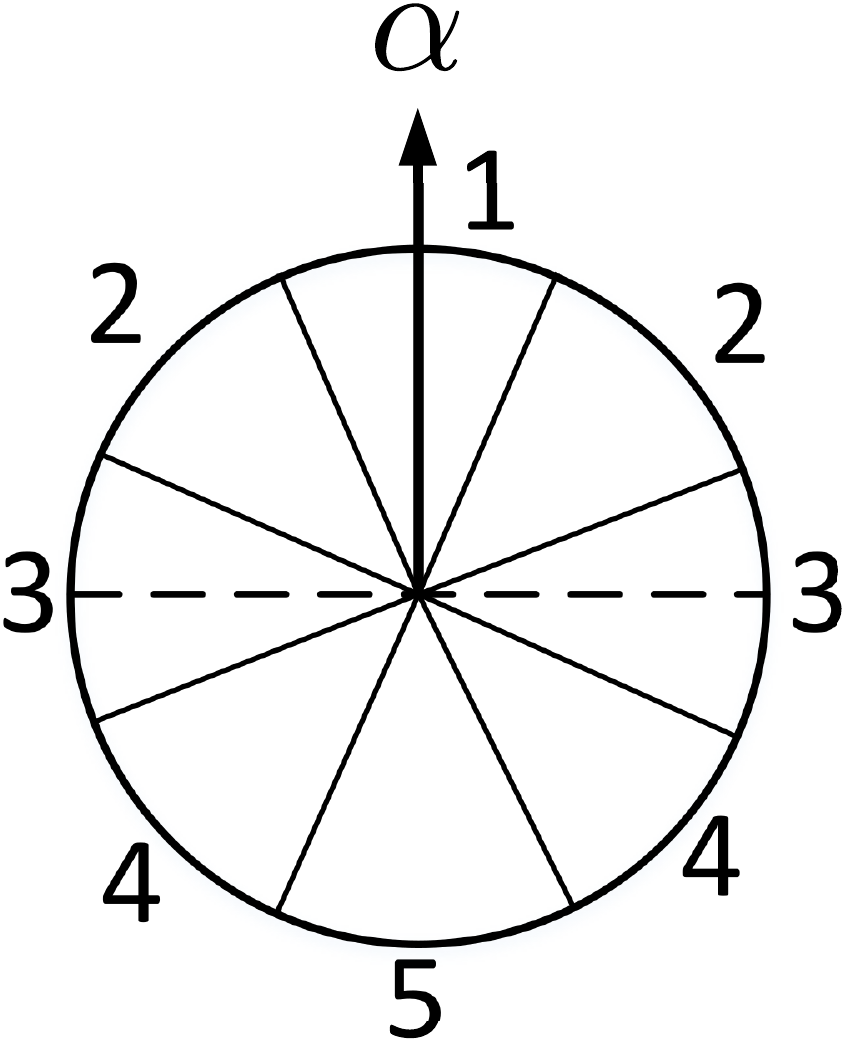}\label{subfig:directionRelation}}
\subfloat[]{\includegraphics[width=0.3\linewidth]{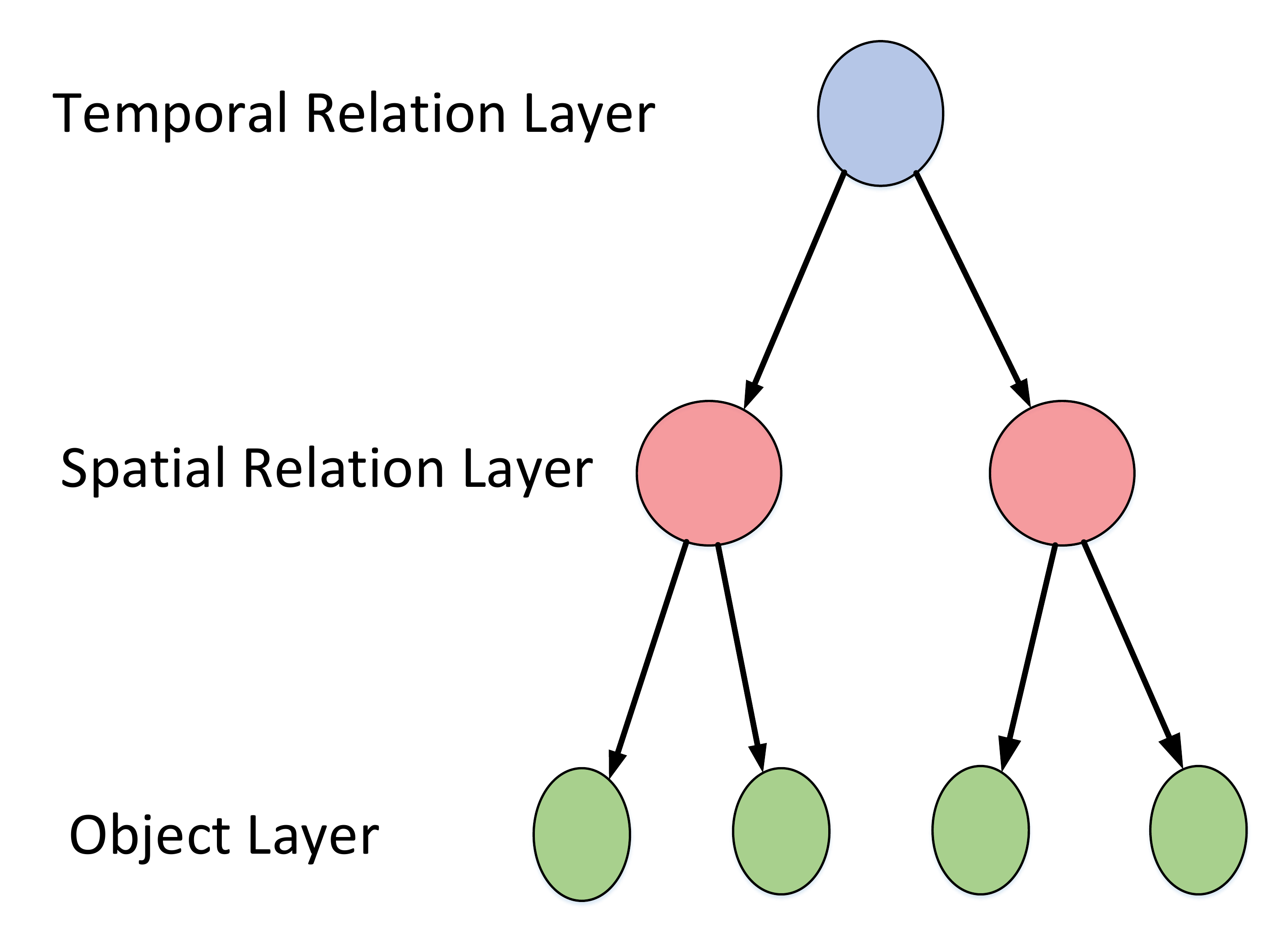}\label{subfig:cellGraph}}
\subfloat[]{\includegraphics[width=0.43\linewidth]{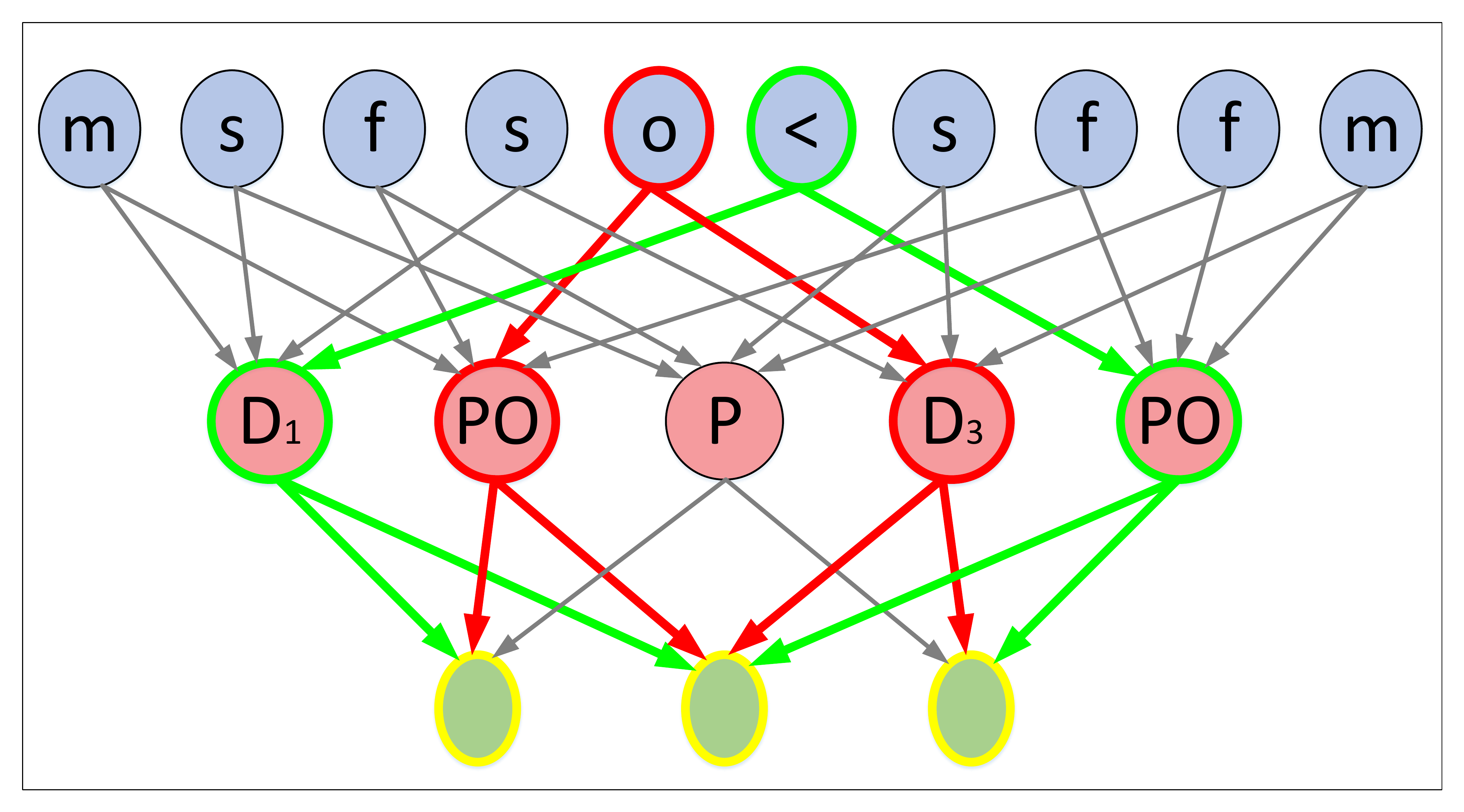}\label{subfig:spatioTemporalGraph}}
\caption{Illustration of basic elements of the directional
qualitative spatio-temporal graph \ref{subfig:spatioTemporalGraph}.
The spatial relations consist of qualitative distance relations
\ref{subfig:distanceRelation} and qualitative direction relations
\ref{subfig:directionRelation}. \ref{subfig:temporalRelation}
illustrates the temporal relations. The cell graph
\ref{subfig:cellGraph} is a hierarchical composition of the spatial
and temporal relations. The cell graphs over a duration form the
qualitative spatio-temporal graph \ref{subfig:spatioTemporalGraph}.}
\label{fig:spatialRelation}
\end{figure*}

\subsection{Basic Definitions}
\subsubsection{Qualitative Distance Relations}
A qualitative distance relation, denoted as $\mathcal{D}_s$, provides an appropriate generalization for the accurate quantitative distance between pairwise objects, as illustrated in Fig. \ref{subfig:distanceRelation}. Specifically, a sequence of detected regions covered by rectangles in 2D frame images is tracked for every participant. For any two objects $O_{i}$ and $O_{j}$, the distance relation belongs to one of three qualitative distance relations, $D(\text{discrete})$, $PO(\text{partial overlap})$, or $P(\text{part, part inverse, and equality})$, based on the proportion of overlap of the two corresponding rectangles, which is formulated as
\begin{equation*}
\frac{A_{O_i} \cap A_{O_j}}{\min(A_{O_i},A_{O_j})},
\end{equation*}
where $A_{O_{k}}$ is the area covered by object $O_{k}$, the numerator is the overlapping area of the two objects, and the denominator is the area of the smaller one. All three relations are symmetric. Qualitative distance relations describe the spatial relations from the perspective of distance, and are affected by the size of relative objects.

\subsubsection{Qualitative Direction Relations}
A qualitative direction relation $\mathcal{D}_r$ is derived by partitioning the 2D space into $n$ bins (similar to Xia et al. \cite{xia2012view}). For robustness we take $n=5$ as shown in Fig. \ref{subfig:directionRelation}. Thus, the inclination angle is divided into five bins from the zenith vector $\alpha$ with a resolution 45\degree. The main purpose of a qualitative direction relation is to reveal the azimuth angular relationship between two objects, e.g., one object is to the upper left of the other one. We emphasize vertical instead of horizontal relations, so upper-left and upper-right direction relations are allocated to the same bin. The allocation is based on the position of the tracking centers of the objects. Direction relations help improve the comprehensiveness of qualitative spatial relations, as demonstrated in our experiments.

\subsubsection{Qualitative Spatial Relations}
A qualitative spatial relation is defined as the complete spatial position relationship between any two objects and is based on their qualitative distance and direction relations. Because the direction relations become susceptible to noise when two objects approach each other, we neglect the direction relations when the distance relations are localized at $PO$ or $P$. As a consequence, we collect seven kinds of qualitative spatial relations: $PO$, $P$, $\{D_i\}_{i=1}^5$, for two objects in 2D space, where $i$ indicates the index of the direction relation. A qualitative spatial relation for pairwise objects is represented as
\begin{equation*}
\mathcal{S}=(\mathcal{D}_s,\mathcal{D}_r),
\end{equation*}
A qualitative spatial relation combines both aspects of distance and direction and is a comprehensive and robust description of spatial relations for any two component objects.

\subsubsection{Qualitative Temporal Relations}
A qualitative temporal relation links two qualitative spatial relations to represent their temporal relationship for a specific time period, and these relations are designed to reflect the variation in human actions over time. We keep the nature of a qualitative description to ensure robustness, and Allen's interval algebra \cite{allen1983maintaining} , which is a calculus for temporal reasoning, is applied to obtain the relative temporal relations, as illustrated in Fig. \ref{subfig:temporalRelation}. Possible relations between time intervals are defined in Allen’s interval algebra, and a composition table is provided that can be used as a basis for reasoning about temporal descriptions of events. All seven qualitative temporal relations except the relation $X=Y$ are asymmetric. In this paper, the qualitative temporal relations $X~s~Y$, $X~d~Y$ and $X~f~Y$ are mapped to one relation for simplicity and compactness. Qualitative temporal relations are an important supplement to qualitative spatial relations in action description.

\subsubsection{Cell Graphs}
A cell graph is a hierarchical model that combines descriptions of qualitative spatial and temporal relations (as stated above) for specific participating objects in a reasonable manner. As shown in Fig.~\ref{subfig:cellGraph}, a cell graph is structured as a complete binary tree, where the vertices are partitioned into three layers and where each layer refers to a different node type. Specifically, the nodes in the object layer that are not explicitly marked with labels correspond to objects in $O$. The nodes in layer 2 represent qualitative spatial relations between pairwise objects, which are represented by nodes in layer 1, and are labeled with one of seven qualitative spatial relations. Likewise, nodes in layer 3 are labeled as one of the five qualitative temporal relations between the respective pairwise nodes in layer 2. The key characteristic of the cell graph is that the hierarchical description is a complete binary tree in which the qualitative spatio–temporal relations hold among objects but the locations or intervals are not represented metrically. Human actions are expressed as complicated graphs comprising several cell graphs. This facilitates the similarity measurement of any two human actions using the graphs. In addition, the cell graph incorporates both the qualitative spatial and temporal relations of component objects and constitutes the basis of the spatio–temporal graph representation for human actions.

\subsection{Graph Construction}
The construction of our improved qualitative spatio-temporal graph is close to that proposed in Sridhar et al. \cite{sridhar2010unsupervised}. We give a brief description of the building process in the following.

Given a video, for any two objects $O_{i}$ and $O_{j}$, a sequence of coordinates in 2D frame images are tracked. Each object is allocated with a neighboring region whose location is based on track coordinates. Note that the region size depends on the corresponding object size, and the human body part object region size is determined by the human body size and the category of the body part. For each frame, the qualitative distance relation $\mathcal{D}_s(O_{if},O_{jf})$ is captured by the proportion of overlap of the corresponding rectangles, and the qualitative direction relation $\mathcal{D}_r(O_{if},O_{jf})$ is computed by the two coordinates according to our basic definitions, where superscript $f$ indicates the frame number. The qualitative distance relation $\mathcal{D}_s$ and direction relations $\mathcal{D}_r$ are computed for every pair of objects participating in the human activity video. A low-pass filter suppresses any jitters resulting from objects and skeleton detection errors in both relations. Given a video sequence, we can obtain a sequence of  qualitative distance and direction relations in this way.

The qualitative spatial relations $\mathcal{S}=\{PO,P,\{D_i\}_{i=1}^5\}$ are found for all pairwise objects in each frame image. As a result, we can arrange these spatial relations in multiple rows. Each row indicates the index of combination for any two component objects and column indicates video frame number. In other words, $\mathcal{S}(O_{if},O_{jf})$ denotes the qualitative spatial relation between objects $i$ and $j$ in the $f$-th frame. Because the spatial relations of some objects change slowly during the activities, each row contains only a few distinctive spatial relations. For convenience, we compress sub-sequences with the same spatial relations in each row and record them with $\mathcal{S}_{ij}(s_t,e_t)$, where $s_t$ and $e_t$ are the start and end frame indexes for $t$-th relation between objects $i$ and $j$. Next, a specific qualitative temporal relation belonging to one of the defined relation types is assigned to two qualitative spatial relations indicated by $\mathcal{S}_{i_{1}j_{1}}(s_{t_{m}},e_{t_{m}})$ and $\mathcal{S}_{i_{2}j_{2}}(s_{t_{n}},e_{t_{n}})$. A qualitative temporal relation linking two qualitative spatial relations that are related to pairwise objects produces a cell graph. The spatio–temporal relations among objects in a specific time period and the dynamic information of the interactions are captured by all derived cell graphs. The whole graph for a video sequence is constructed over time and contains every qualitative temporal relation of the mutual qualitative spatial relations generated by the pairwise objects. Fig. \ref{subfig:spatioTemporalGraph} shows an example of the constructed qualitative spatio-temporal graph where the corresponding action is performed by three objects.

A spatio–temporal graph constructed for a human activity video sequence is an intuitive collection of basic cell graphs. Any qualitative spatial and temporal interactions among participating objects are captured by the graph. Moreover, the spatio–temporal graph is a comprehensive representation for a video event, especially for a short-term video event like a human action.

\section{Human Action Representation}
\label{sec:representation}

To recognize human activity, the first step is to represent human action reliably. In this section, we give a detailed description of human action representation based on the above qualitative spatio–temporal graph. In particular, a long human activity is divided into several successive human actions with sliding windows to first obtain long-term dynamic information. Then, a hierarchical decomposition of the human body is utilized to construct the local spatio–temporal graph. Finally, histogram-based statistics of the human actions are employed to represent the human activity.

\subsection{Capturing Dynamic Patterns with a Sliding Window}
A qualitative spatio–temporal graph is a suitable model for short-term objects interactions, however, the graph only captures local temporal relations, and it is not able to capture the long-term dynamic information that reflects the evolution of object interactions for a human activity. Therefore, a single qualitative spatio–temporal graph is not a suitable representation for an entire human activity video. Instead, sliding windows are employed to cover the sub-sequences of a video in which multiple object interactions arise, and then qualitative spatio–temporal graphs are constructed from the actions in this window to obtain the corresponding spatio–temporal patterns. In this way, we can represent local action with information in the corresponding window and capture the dynamics of activity evolution with the transformations between adjacent windows.

Most existing sliding-window methods fix the window length and step length according to the number of frames, but this leads to meaningless and repetitive qualitative spatio–temporal graphs when no spatial relation changes occur. In our method, we divide the video frames into fragments when there is any change in the qualitative spatial relations. Fragments are basic units in which the same spatial relations for all objects hold over the whole period. As a result, the window and step lengths are based on the number of fragments rather than the number of frames. In particular, a window of length $l_{w}$ spans the relevant $l_{w}$ fragments, which includes $l_{w} - 1$ times spatial relation changes, and we jump over $l_{s}$ (step length) fragments every time step.

For a human activity video, the qualitative spatial relations for every component pairwise object are computed according to the above definition in each frame, and a sliding window based on spatial relations makes the method practical when an activity is divided into several successive potential actions. Then, time series models can be trained and used to predict the activity label.

\subsection{Hierarchical Decomposition of Human Body}
It is obvious that two relations can be unequal even though they have the same qualitative spatio–temporal graph description (left/right feet and torso or left/right hands and head). However, the identities of the component objects are ignored during the construction of a qualitative spatio–temporal graph, which results in ambiguity between two spatio–temporal relations generated from different objects. However, adding an identity label to the participating objects would lead to the exponential growth of the number of basic units (cell graphs) and make the qualitative spatio–temporal graph feature sparse. Therefore, a balanced hierarchical decomposition of human body is employed in our method.

\begin{figure}[t]
\centering
\includegraphics[width=0.9\linewidth]{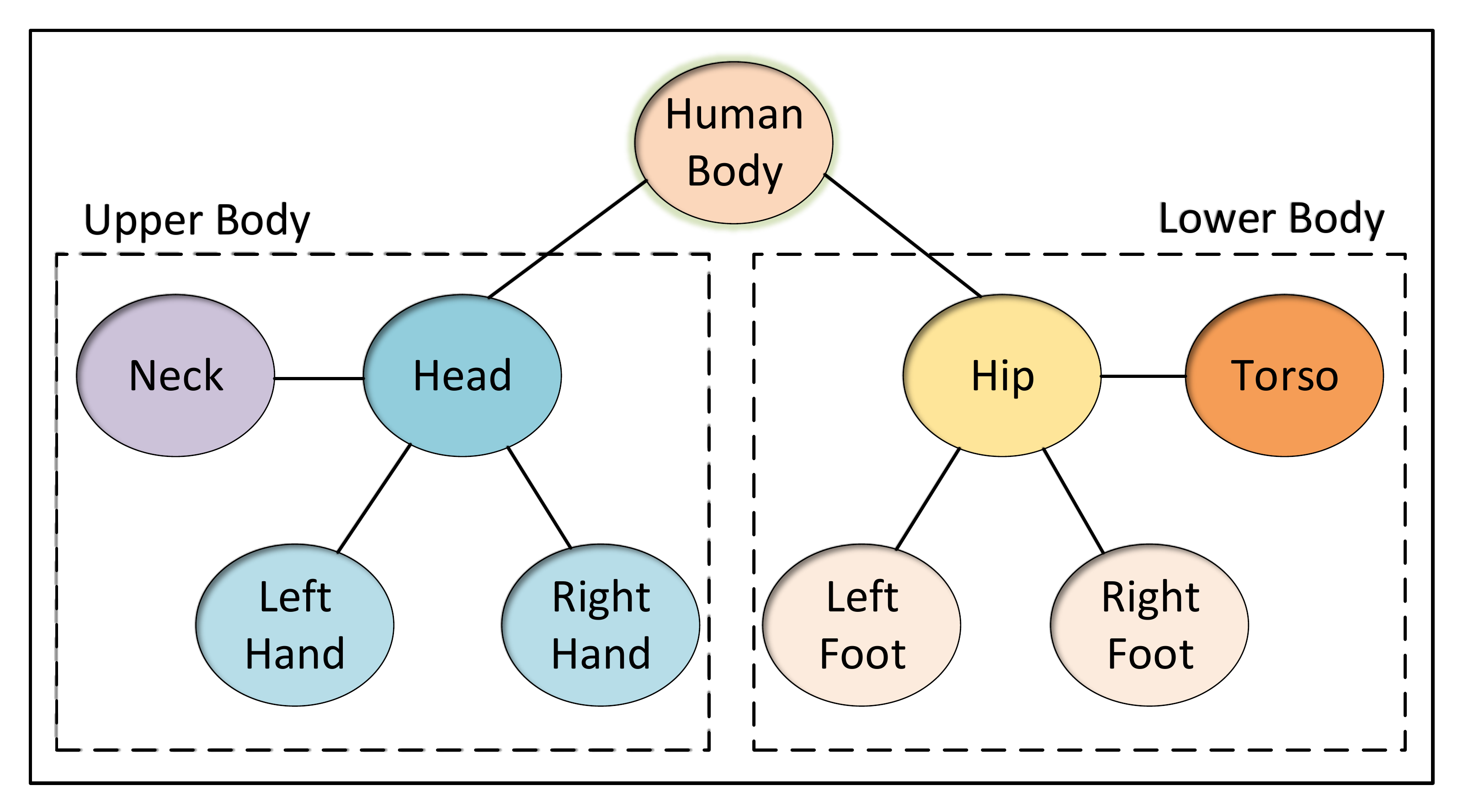}
\caption{Hierarchical decomposition of the human body. The upper body part comprises the head, neck, and left/right hands. The lower body part comprises the hip, torso, and left/right feet.}
\label{fig:hhd}
\end{figure}

The fundamental scheme of hierarchical decomposition of human body is shown in Fig. \ref{fig:hhd}, in which the human body is partitioned into two main parts: the upper body (including the head, neck, and left/right hands) and the lower body (including the hip, torso, and left/right feet). This decomposition efficiently distinguishes actions that are performed by different parts of the human body. More importantly, the qualitative spatio–temporal graph constructed from the whole human body captures the global features of a human action, whereas the upper body and lower body parts capture local features. These two kinds of information supplement each other and improve the discrimination of qualitative spatio–temporal graph representation for human action.

Here, we use the random forest-based method in \cite{shotton2011real} to predict the segmentation of a human body into parts from a single depth image, where the parts defined to be spatially localized near skeletal joints of interest. The complexity of the posture recognition is $O(M \cdot mtry \cdot Nlog(N))$, where $M$ is the number of trees in the random forest, $mtry$ is the number of the sampled variables at each node, and $N$ is the number of the nodes to train in the random forest.

In the implementation, in addition to constructing a single qualitative spatio–temporal graph for the whole human body and participating objects, graphs with the same form are constructed for both the upper body and lower body parts, which display more concrete interactions with objects. The representation of specific human actions are obtained by concatenating all the qualitative spatio–temporal graphs constructed for different decompositions of parts during the same time period.

\subsection{Histogram Statistics of Spatio-Temporal Graph}
To easily measure the difference between human activities, the derived qualitative spatio–temporal graphs are transformed into feature vectors. A spatio–temporal graph represents the interactions among objects during a specific period of time, and it displays both the qualitative spatial and temporal relations. However, there are no existing methods for learning information from such graphs directly; thus, a key procedure is to convert the graph we constructed from a sequence of video into a measurable feature vector. In this way, a potential action can be represented by an observation symbol.

The qualitative spatio-temporal graph is an assembly of multiple cell graphs, which are built by seven qualitative spatial relations and five qualitative temporal relations, as shown in Fig. \ref{subfig:spatioTemporalGraph}. A cell graph is a combination of qualitative temporal relation nodes linking two qualitative spatial relation nodes whereas every qualitative spatial relation node links two indiscriminate object nodes. Because of the asymmetry of the qualitative temporal relations (except for relation $X$ $=$ $Y$), the number of types of cell graph $N_{cg}$ according to permutation and combination formulae is,
\begin{equation*}
N_{cg} = N_{s}^2 \cdot N_{at} + N_{s} \cdot N_{st}
\end{equation*}
where $N_{s}$ is the number of qualitative spatial relation types and $N_{st}$ and $N_{at}$ are the numbers of symmetric and asymmetric qualitative temporal relations types, respectively.

We build a cell graph dictionary $\{cg_{1},cg_{2},\cdots,cg_{N_{cg}}\}$ such that any qualitative spatio-temporal graph can be expressed in terms of a bag of cell graphs (BoCG). Then a histogram is constructed that places all cell graphs into corresponding bins. A qualitative spatio-temporal graph $g_{i}$ is represented as a histogram of length $N_{cg}$,
\begin{equation}
\phi(g_{i}) = [f_{i1},f_{i2},\cdots,f_{iN_{cg}}],
\end{equation}
where $f_{ij}$ is the frequency at which a specific cell graph $cg_{j}$ appears in the corresponding qualitative spatio-temporal graph $g_{i}$. To compare any two qualitative spatio-temporal graphs (potential actions), the BoCG kernel $\mathcal{K}$ is constructed in terms of the cell graph dictionary as follows:
\begin{equation}
\mathcal{K}_{ij} = <\phi(g_{i}),\phi(g_{j})> = \sum_{k=1}^{N_{cg}}{f_{ik}f_{jk}}.
\end{equation}
The BoCG kernel provides a measure for the qualitative spatio–temporal graph and is used to quantize and symbolize the feature vector.

\section{Discrete HMMs for Activity Recognition}
\label{sec:recognition}
We adopt discrete HMMs to model the evolution of human actions because of their ability to process dynamic information, as demonstrated in some previous work \cite{rabiner1989tutorial},\cite{xia2012view}. All of the steps of this process as well as the background knowledge are discussed here to make this paper self-contained. The following subsections first state the process of vector quantization to obtain discrete symbol sequences, which are then applied to human activity  recognition with discrete HMMs.

\subsection{Vector Quantization}
Each human activity is represented by a human action sequence, which consists of a series of feature vectors. Our aim is to convert each human action feature vector into an observation symbol such that human activity is denoted by a sequence of observation symbols. Note that each qualitative spatio–temporal graph described in the previous section is indicated by a high dimensional vector in a continuous space. To reduce the number of observation symbols, vector quantization is performed to cluster the feature vectors. However, many common human actions are shared by most human activities whereas significant and distinguishing human actions are fewer in number; this imbalance leads to inferior vector quantization results when we apply K-means to cluster all the data directly. To resolve this problem, we collect a sufficient number of distinct human actions and compute their qualitative spatio–temporal graph vectors. The collected vectors are clustered into $K$ clusters (forming a $K$-word vocabulary) using $K$-means, then all vectors are assigned to one of the $K$ clusters. As a consequence, one vector is treated as a single symbol of a visual word, and each human activity is represented by a sequence of these symbols.

\subsection{Activity Recognition with Discrete HMMs}
The discrete HMMs are trained based on the obtained visual word sequences for human activity recognition. In discrete HMMs, sequential data are treated as the output of a Markov process whose state cannot be directly observed. As previously discussed, each divided human activity sequence is assigned to a series of visual words and used as the inputs to train discrete HMMs, which are then utilized to predict for unknown human activities in a video.

An HMM that has $N$ states $S = \{s_{1},s_{2},...,s_{N}\}$ and $M$ observed symbols $O = \{o_{1},o_{2},...,o_{M}\}$ can be formulated as $H = \{A,B,\pi\}$. The state transition matrix $A\in\mathbb{R}^{N \times N}$ is,
\begin{equation}
A = \{a_{ij}|a_{ij} = P(s_{t+1}=q_{j}|s_{t}=q_{i})\},
\end{equation}
where the state at time step $t$ is indicated as $s_{t}$, $a_{ij}$ denotes the transition probability from the state $q_{i}$ to $q_{j}$, and the confusion matrix $B\in\mathbb{R}^{N \times M}$ is,
\begin{equation}
B = \{b_{ik}|b_{ik} = P(o_{t} = p_{k}|s_{t} = q_{i})\},
\end{equation}
where $b_{ik}$ denotes the observation probability of $p_{k}$ given $q_{i}$ at specific time step $t$. Finally, the initial state distribution $\pi$ is,
\begin{equation}
\pi = \{\pi_{i}|\pi_{i} = P(s_{1} = q_{i})\}.
\end{equation}

For each class of human activity, an HMM is trained based on the observation sequences of the activities. Discrete HMMs are the probability distributions over the observation sequences of a specific human activity. For an unknown human activity video, an observation symbol sequence $O = \{O_{1},O_{2},\cdots,O_{T}\}$ derived from our general framework is fed into each model $\mathcal{H}$ to compute its probability $P(O|\mathcal{H})$, which can be solved with the forward algorithm. The label $l$ of the human activity is determined by the model class that obtains the highest likelihood, that is,

\begin{equation}
\text{label} \quad l = \arg\max\limits_{i\in\{1,2,\cdots,C\}} \{P(O|\mathcal{H}_{i})\},
\end{equation}
where $P(O|\mathcal{H}_{i})$ is the likelihood on the $i$-th HMM and $C$ is the total number of human activity classes. This model is able to capture sufficient dynamic information for distinguishing different human activities.

The time complexity of the discrete HMM is $O(N^2 \cdot T)$, where $N$ is the number of the states in the discrete HMM and $T$ is the length of the input sequence.

\section{Experimental Studies}
\label{sec:experiment}
\begin{figure*}
\centering
\includegraphics[width=\linewidth]{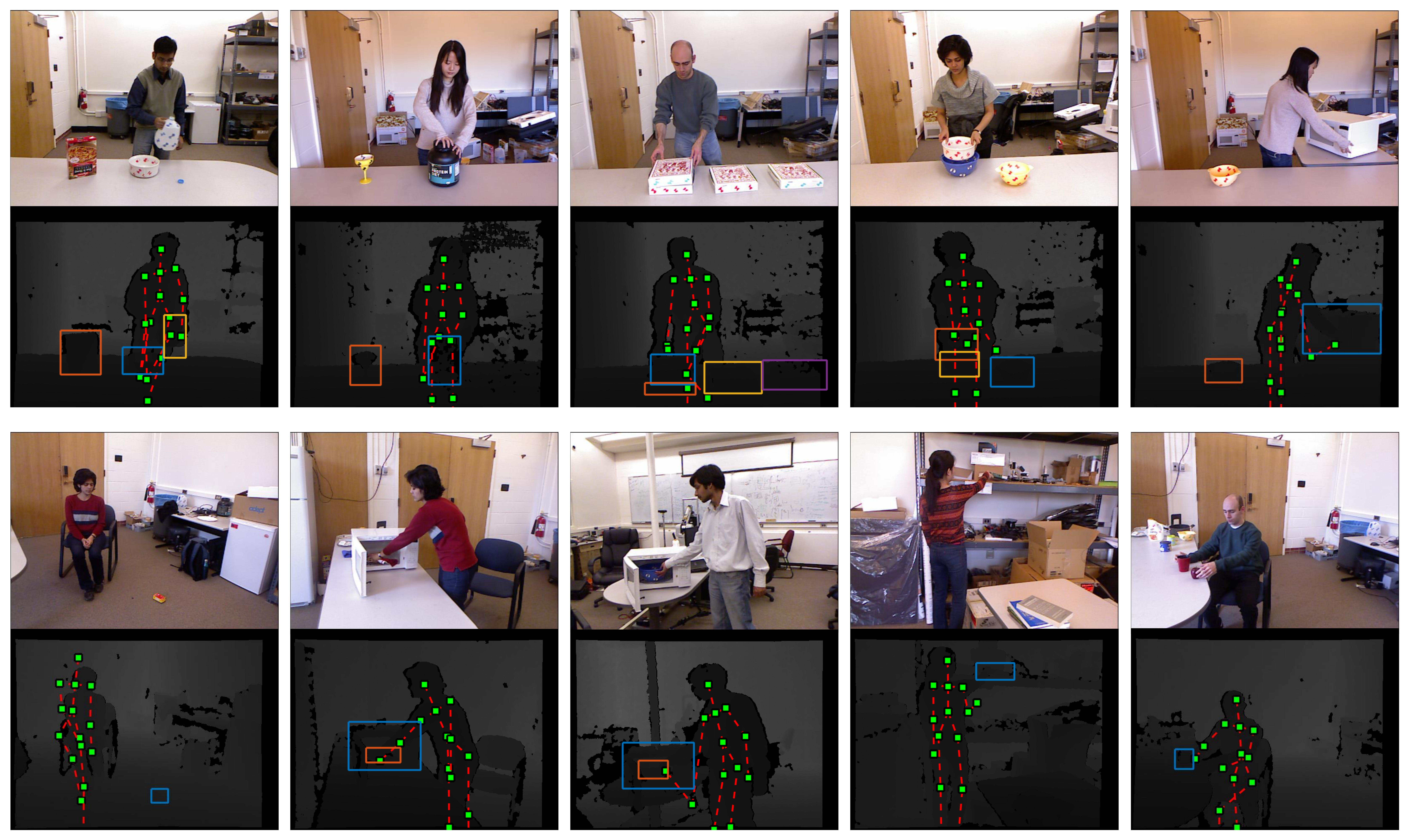}
\caption{Illustration of the sample images from videos of the 10 activities in the CAD-120 dataset. We show the RGB image frames as well as the corresponding depth maps (only depth images are used in the experiments), where 2D skeletal information and object positions are annotated in green and other colors, respectively. The type of activity from left to right and top to bottom is $\mathit{Making}$ $\mathit{Creal}$, $\mathit{Taking}$ $\mathit{Medcine}$, $\mathit{Stacking}$ $\mathit{Objects}$, $\mathit{Unstacking}$ $\mathit{Objects}$, $\mathit{Microwaving}$ $\mathit{Food}$, $\mathit{Picking}$ $\mathit{Objects}$, $\mathit{Cleaning}$ $\mathit{Objects}$, $\mathit{Taking}$ $\mathit{Food}$, $\mathit{Arranging}$ $\mathit{Objects}$, and $\mathit{Having}$ $\mathit{a}$ $\mathit{Meal}$, respectively.}
\label{fig:cad120samples}
\end{figure*}

To evaluate the proposed dynamic spatio–temporal method for human activity recognition, experiments were performed on the CAD-120 3D activity benchmark datasets. We first detail the experimental implementation, which is followed by a description of the experimental dataset. Then we present the experimental results. Finally, we present an ablation analysis of the proposed approach.

\subsection{Implementation Details}
The launch of Kinect led to  a low-cost and real-time solution for the estimation of the 3D locations of objects or persons in a scene. In \cite{shotton2013real}, the authors extracted 3D body joint locations from a depth image using an object recognition scheme and labeled human body parts based on per-pixel classification results. The parts include the LU/RU/LW/RW head, neck, L/R shoulder, LU/RU/LW/RW arm, L/R elbow, L/R wrist, L/R hand, LU/RU/LW/RW torso, LU/RU/LW/RW leg, L/R knee, L/R ankle, and L/R foot (where L = left, R = right, U = upper, and D = down). In our experiments, we obtained the 2D locations of 15 skeletal joints including the head, neck, torso, L/R shoulders, L/R elbows, L/R hips, L/R knees, L/R hands, and L/R feet using this method. Some sample RGB frame images with their corresponding depth images and colored skeleton joint information are shown in Fig.~\ref{fig:cad120samples}. The 2D skeletal joints are shown in green and the participating objects we used in our experiment are indicated with other colors.

In the proposed method, the qualitative direction relations are computed directly from the acquired 2D locations, whereas the qualitative distance relations are determined by the proportion of overlap between two corresponding rectangles. There are obvious variations in the size of the human body parts due to their original sizes and the variations in distance between the camera and human body. To determine the size of the relative rectangle that covers the area occupied by an object in 2D space, we take the horizontal distance between the left hip and right hip as the basic rectangle length $l_{b}$ and the vertical distance between the neck and torso as the basic rectangle width $w_{b}$ for each subject. As a result, the length and width of all rectangles representing the parts of human body are multiples of $l_{b}$ and $w_{b}$, respectively.

The qualitative temporal relations were obtained according to the variations in pairwise qualitative spatial relations. Based on this information, we constructed the relative qualitative spatio–temporal graph for every part of the human body (the upper and lower human body) with the participating objects over specific time periods determined by the width of the sliding window and step length. Histogram statistics for all generated graphs were then concatenated. To improve convergence, we only collected a sufficient number of distinct vectors and used K-means to reduce the number of observation symbols. In this way, the derived human action sequences were converted to a series of symbols for both training and testing human activity videos. Finally, sequences of symbols that represent spatio–temporal relations were used to train the discrete HMMs and predict the label of unknown activities based on the trained model.

\subsection{CAD-120}

CAD-120 \cite{koppula2013learningb} is the human activity dataset comprising 120 RGB-D human activity video sequences with skeleton information (the dataset can be downloaded from http://pr.cs.cornell.edu/humanactivities/data.php). These sequences represent different daily living activities which were performed by two males and two females. Each participant was given one of 10 high-level human activities to perform, and each person performed each high-level activity three times. In addition, the human body parts were recorded using a Microsoft Kinect camera. Among the four subjects, three were right handed and one was left handed. Each human activity video is labeled with a single high-level activity: $\mathit{Making}$ $\mathit{Cereal}$, $\mathit{Taking}$ $\mathit{Medicine}$, $\mathit{Stacking}$ $\mathit{Objects}$, $\mathit{Unstacking}$ $\mathit{Objects}$, $\mathit{Microwaving}$ $\mathit{Food}$, $\mathit{Picking}$ $\mathit{Objects}$, $\mathit{Cleaning}$ $\mathit{Objects}$, $\mathit{Taking}$ $\mathit{Food}$, $\mathit{Arranging}$ $\mathit{Objects}$, and $\mathit{Having}$ $\mathit{a}$ $\mathit{Meal}$. Both automatic and ground truth tracks of the participating objects are presented in each video, although the automatic track is noisy due to the restrictions of the tracking algorithm. Frame images sampled from videos of the 10 classes of human activities in the CAD-120 dataset are shown in Fig.~\ref{fig:cad120samples}.

As our proposed method relies on the skeleton information to construct qualitative spatio-temporal graphs, some famous datasets for human action recognition such as Charades cannot be adopted to test our proposed method due to the lack of the skeleton information in these datasets. Compared to other datasets with the skeleton information such as Cornell Activity Dataset 60 (CAD-60) and Daily Activity 3D Dataset, CAD-120 is more appropriate for evaluating our proposed method. This is because that CAD-120 has two unique advantages to evaluate our method. First, every human activity in the video sequences in CAD-120 consists of a series of successive human actions, which enables the validation of our approach for modeling the evolution of these actions. Second, each human action in the video sequences in CAD-120 involves an interaction of a person and at least one object, which helps to validate the capability of our proposed method to capture significantly distinguishing features from the interactions.

\subsection{Experimental Results}
We follow the same train-test split as the one implemented in \cite{koppula2013learningb} for our evaluation and comparison. We also used 4-fold cross validation approach is applied, in which three subjects are used for training and the fourth new subject is used for testing. One HMM $\mathcal{H}$ is trained for one high-level class using the human activity video sequences in the same category on three training subjects, and then the trained HMMs are utilized to predict the label of the unknown activities performed by the fourth subject.
\begin{table*}[!t]
\renewcommand{\arraystretch}{1.4}
\centering
\caption{Performance measurements on the CAD-120 dataset in comparison with or without ground-truth temporal segmentation based on accuracy, precision, and recall.}
\label{tab:results}
\begin{tabular}{p{3.9cm}p{2.5cm}p{2.7cm}p{2.5cm}}
\toprule
Methods & Accuracy & Precision & Recall \\ \hline
\multicolumn{4}{c} {\emph{With ground-truth temporal segmentation}} \\ \hline
Koppula et al.\cite{koppula2013learningb} & $84.7 \pm 2.4$ & $85.3 \pm 2.0$ & $84.2 \pm 2.5$ \\
Koppula and Saxena \cite{koppula2013learninga} & $93.5 \pm 3.0$ & $95.0 \pm 2.3$ & $93.3 \pm 3.1$ \\ \hline
\multicolumn{4}{c} {\emph{Without ground-truth temporal segmentation}} \\ \hline
Koppula et al.\cite{koppula2013learningb} & $80.6 \pm 1.1$ & $81.8 \pm 2.2$ & $80.0 \pm 1.2$ \\
Koppula and Saxena \cite{koppula2013learninga} & $83.1 \pm 3.0$ & $87.0 \pm 1.6$ & $82.7 \pm 3.1$ \\
Tayyub et al. \cite{tayyub2014qualitative} & $\bm{95.2} \pm 2.0$ & $\bm{95.2} \pm 1.6$ & $\bm{95.0} \pm 1.8$ \\
DSTR-gt-tracks & $93.3 \pm 2.3$ & $94.7 \pm 2.5$ & $94.0 \pm 2.4$ \\ \hline
\multicolumn{4}{c} {\emph{With ground-truth temporal segmentation and without ground-truth object bounding boxes}}  \\ \hline
Koppula et al. \cite{koppula2013learningb} & $75.0 \pm 4.5$ & $75.8 \pm 4.4$ & $74.2 \pm 4.6$ \\
Rybok et al. \cite{rybok2014important} & 78.2 & - & - \\
Tayyub et al. \cite{tayyub2014qualitative} & $75.8 \pm 6.8$ & $77.9 \pm 11.0$ & $75.4 \pm 9.1$ \\
DSTR-auto-tracks & $\bm{89.2} \pm 2.5$ & $\bm{90.2} \pm 3.4$ & $\bm{89.9} \pm 2.5$ \\
\bottomrule
\end{tabular}
\end{table*}

Because of the randomness existing in the initialization of both cluster centers and the discrete HMM algorithm, we ran the experiment 30 times and report the mean performance. The number of clusters and states were set to $K = 38$ and $N = 7$, respectively. A comparative summary of the performance of previous approaches and our developed dynamic spatio-temporal relations (DSTR) method is shown in Table \ref{tab:results}. The proposed method achieves a competitive results with an accuracy of $93.3\%$, a precision of $94.7\%$, and a recall of $94.0\%$. Compared to benchmark methods, which require the ground-truth temporal segmentation of the sub-level activities, our method is able to obtain equivalent results without this additional information. Moreover, when no ground-truth temporal segmentation of the sub-activities is provided, the proposed method shows a substantial improvement of $10.1\%$, $7.7\%$, and $11.3\%$ in terms of accuracy, precision, and recall.

Compared to the highest performance of \cite{tayyub2014qualitative}, which combines quantitative spatial, qualitative spatial and temporal features, feature selection, and SVM(Support Vector Machine), our results are only $1.9\%$, $0.5\%$, $1.0\%$ lower in accuracy, precision, and recall, respectively. As \cite{tayyub2014qualitative} demonstrated,the quantitative spatial relations form the dominant feature for classification. Because our approach only requires qualitative representations, this indicates it has more robust performance. The DSTR-gt-tracks results presented in \ref{tab:results}  are based on ground-truth object tracks. More realistic scenarios with noisy automatic object track data are provided by the CAD-120 dataset, where the noisy automatic object tracks were obtained using pre-trained object detectors on a set of frames sampled from the video and particle filter tracker. As the results for DSTR-auto-tracks in \ref{tab:results} show, compared to the state-of-the-art results, we achieve a major improvement of $11.0\%$, $12.3\%$, and $14.5\%$ in terms of accuracy, precision, and recall, respectively. These results are comparable to the results obtained using ground-truth object tracks. This demonstrates the good robustness of the proposed approach, which will make it more applicable in real-world scenarios.

\begin{figure}[t]
\centering
\includegraphics[width=0.8\linewidth]{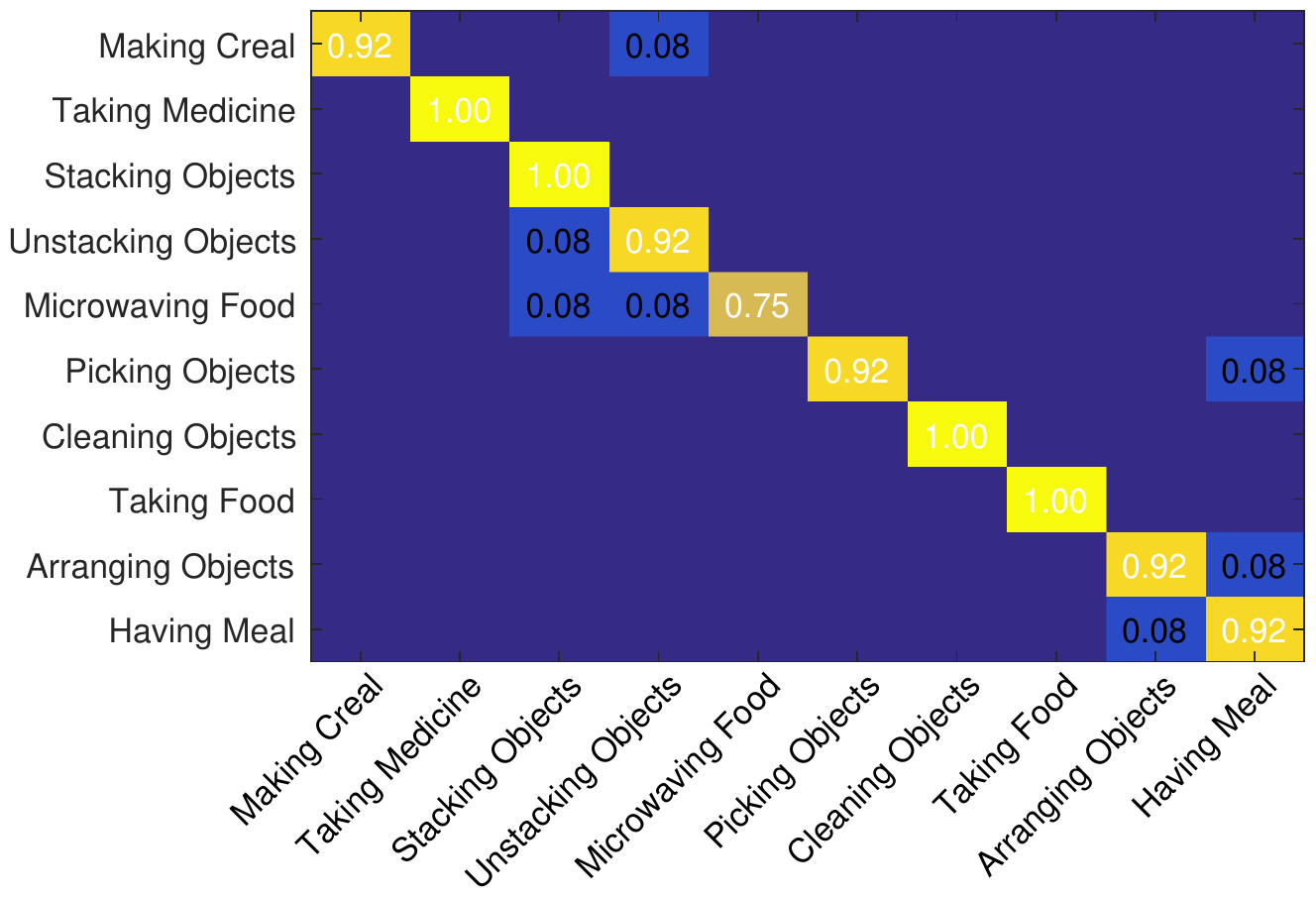}
\caption{Confusion matrix of the proposed method on CAD-120 with ground-truth object tracks on the cross-person setting.}
\label{fig:colorTruthMatrix}
\end{figure}

\begin{figure}[t]
\centering
\includegraphics[width=0.8\linewidth]{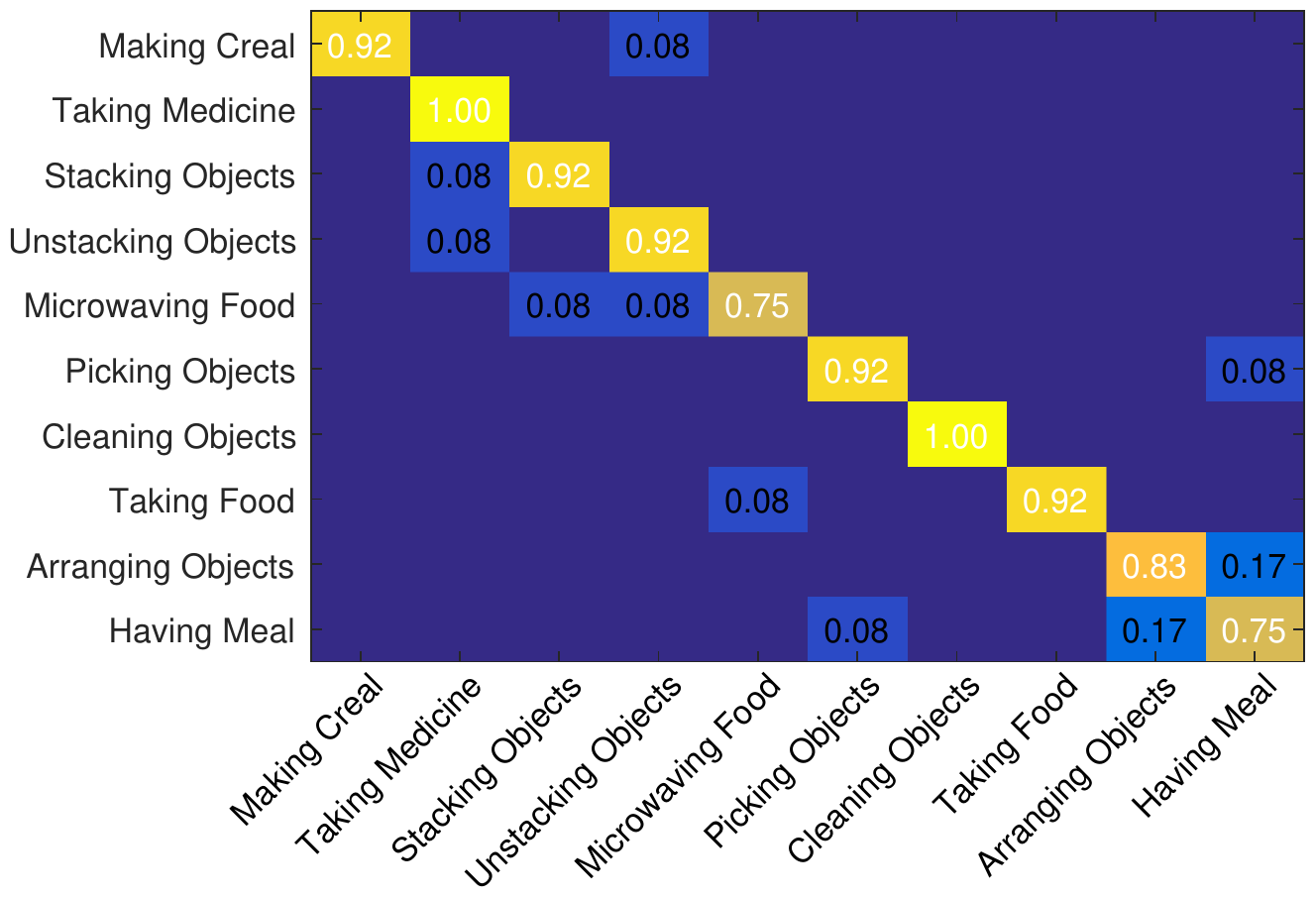}
\caption{Confusion matrix of the proposed method on CAD-120 with automatically extracted object tracking on the cross-person setting.}
\label{fig:colorAutoMatrix}
\end{figure}

In addition, the confusion matrix results of the proposed method with respect to the ground-truth and automatic object tracks are presented in Fig.~\ref{fig:colorTruthMatrix} and Fig.~\ref{fig:colorAutoMatrix}, respectively. The strong diagonals indicate that the proposed method achieves competitive recognition performance on different high-level activities. These results demonstrate both the robustness and discrimination of qualitative spatio–temporal graph representation. Moreover, our approach efficiently and effectively captures the dynamic interaction transitions among participating objects in an activity. However, both confusion matrices show high ambiguity between the human activities $\mathit{Arranging}$ $\mathit{Objects}$ and $\mathit{Having}$ $\mathit{a}$ $\mathit{Meal}$, which is mainly due to the fact that some of the activity videos are almost the same from the perspective of object interactions, even though they belong to different activity classes. To more clearly distinguish these activities, finer-grained information must be  taken into consideration.

\subsection{Ablation Analysis}
We analyze the components of the proposed method to gain insights into the advantages of each. Three components corresponding to our main contributions are analyzed in the following: the dynamic spatio–temporal representation, qualitative direction relations, and hierarchical decomposition of human body. 

\begin{figure}[t]
\centering
\includegraphics[width=\linewidth]{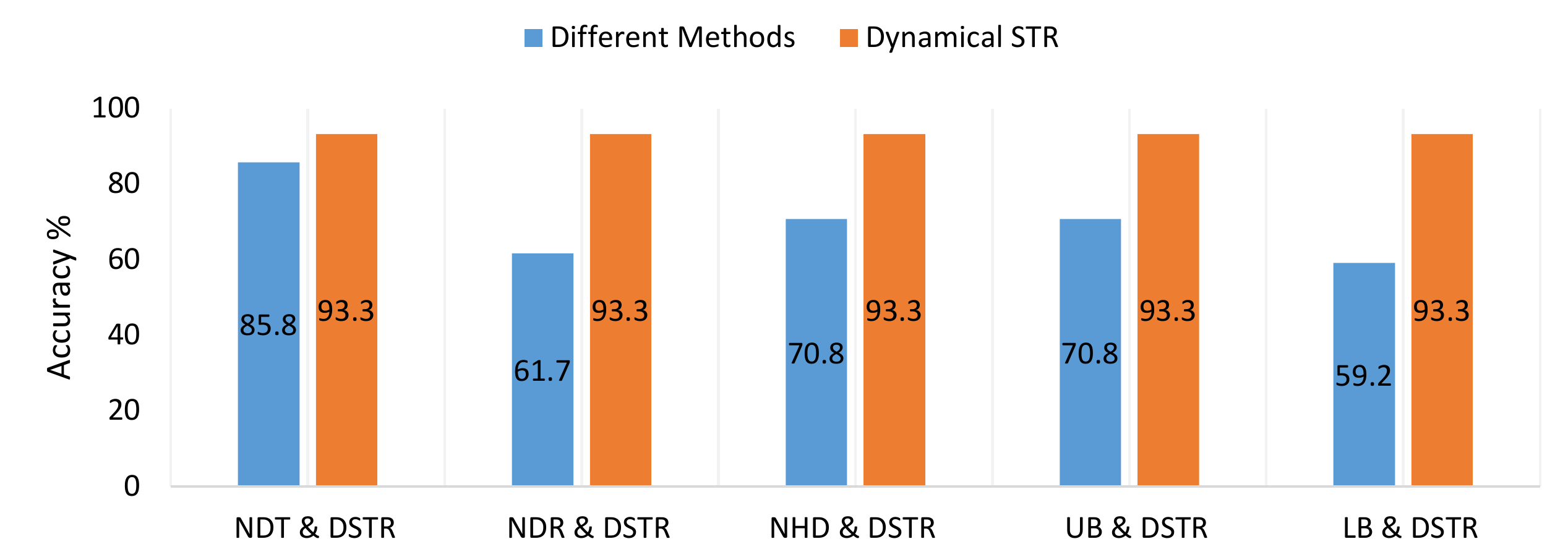}
\caption{Ablation results for the proposed method. DSTR: dynamic spatio–temporal relations, NDT: no dynamic transitions, NDR: no direction relations, NHHD: no human hierarchical decomposition, UB: upper body, and LB: lower body.}
\label{fig:barChart}
\end{figure}

\subsubsection{Dynamic Spatio-Temporal Representation}
First we show the importance of dynamic spatio–temporal representation for activity recognition. We constructed the whole spatio–temporal graph for an entire activity sequence and ignored the dynamical information among successive actions, and then used a simple one-nearest neighbor for classification. The result of no dynamic transitions (NDT) is presented in the first column of Fig.~\ref{fig:barChart}. We find that the accuracy increased by $7.8\%$ when dynamic spatio-temporal representation is used, which reflects the significance of modeling the evolution of human actions (sub-activities). However, the comparative result for NDT of $85.8\%$ still demonstrates the excellent representability of the basic complete spatio-temporal graph for human-object interaction activity.

\subsubsection{Qualitative Direction Relations}
Next, we investigate the strength of improved spatio–temporal
representation (appending qualitative direction relations) versus
prior spatio–temporal representation (using qualitative distance
relations only, i.e., no direction relations (NDR)). We find that
considerable improvement is achieved when qualitative direction
relations are incorporated. The second column in
Fig.~\ref{fig:barChart} shows that including the qualitative
directional relations in the spatio–temporal representation for
activity recognition leads to an improvement of $31.6\%$ in
accuracy. Intuitively, the qualitative direction relations are
robust and complete the spatio–temporal graph representation of an
action.

\subsubsection{Hierarchical Decomposition of Human Body}
Finally, to evaluate the importance of the hierarchical decomposition of the human body for activity recognition, we constructed spatio–temporal graphs for actions based on three alternative decompositions: the whole body (no hierarchical decomposition; NHD), upper body (UB) parts only, and lower body (LB) parts only. The last three columns of Fig.~\ref{fig:barChart} present the comparison results. The evident improvements of $25.5\%$, $25.5\%$, and $34.1\%$ obtained by the full decomposition in terms of accuracy with respect to NHD, UB, and LB indicates that the full hierarchical decomposition of the human body is indispensable for better recognition accuracy.

\section{Conclusion}
\label{sec:conclusion}
In this paper, we developed an efficient method for human activity recognition from the perspective of human action modeling and its dynamical evolution. This method decomposes the problem of human activity recognition into two major aspects. One is to represent a single human action distinctively and robustly, and the other is to model the evolution of human actions regardless of their execution rates. We extended the previously proposed qualitative spatio–temporal graph to include qualitative direction relations to represent human actions and model temporal evolution of human actions with discrete HMMs. In this process, a fully automatic and efficient tool based on a sliding window is used to divide human activity into successive human actions. We further proposed a hierarchical decomposition of the human body into different parts to discriminatively represent distinctive actions. Experimental results on CAD-120 demonstrate the effectiveness of the proposed method, which achieves state-of-the-art results using only object position information. The comparisons with other relevant methods also highlight the importance of the improvements in the proposed method.

The main advantage of the proposed method is the hierarchical abstraction for activity recognition. This abstraction is capable of predicting future human actions according to previous human action symbols, which is the basic idea of sports-game automatic commentary. At present, this approach could be useful in applications such as health care and social assistance, human–computer entertainment, and many other activity recognition fields. However, the limitations of the proposed approach are obvious. First, it is restricted to long-term activities. Because we need to divide an activity into action sequences, the activity video should be long enough to obtain meaningful actions by division. Second, it requires depth data. Although the qualitative spatio–temporal graph is a distinctive and robust representation of human action, its construction requires the information of some important human body joint positions, which can only be captured relatively accurately with depth images for now. These two limitations are the main reason we chose the CAD-120 benchmark to validate the proposed method. On the one hand, this data set provides long-term activities. On the other hand, the human body parts were all recorded using a Microsoft Kinect camera. Finally, the proposed method requires interactions between a human body and objects. We build the spatio–temporal graph based on a fundamental assumption that human activities can be seen as interactions involving two or more persons and/or objects, that is, it is unable to recognize any activities from a stationary video in which no interactions occur. Fortunately, the majority of human activities in the real-world fall under this assumption.

Future work includes meticulously recognizing activities.
\cite{rohrbach2012database} in which the interactions of participating objects are not the only or main factor in human activity recognition. Recognizing these activities is a challenging problem because of the small inter-class distances and large intra-class variations resulting from their diverse subjects and components. A combination of dynamic spatio–temporal relations with specially designed frame image features could be a promising approach that takes both coarse-grained and fine-grained information into consideration. Another direction for future development is the incorporation of the proposed methods with probabilistic models \cite{chen2009probabilistic,chen2013efficient} for probabilistic outputs. The large-scaled probabilistic models \cite{jiang2017scalable} and multi-class classification models \cite{lyu2019multiclass}  will be studied as well.

\bibliographystyle{unsrt}
\bibliography{CIM}

\end{document}